%% file: paper_main.tex
\documentclass[letterpaper, 10 pt, conference]{ieeeconf}  

\IEEEoverridecommandlockouts                              

\usepackage{color}
\usepackage{booktabs}   
\usepackage{multirow} 
\usepackage{amsmath} 
\usepackage{array} 
\usepackage{arydshln}
\usepackage{makecell}
\usepackage{graphicx}
\usepackage{amssymb}
\usepackage{censor}

\title{\LARGE \bf
Repurposing RGB-based Foundation Model for Depth Estimation on \\ Thermal Images Using Hierarchical Supervision
}

\author{Jie Hong$^{1}$, Tingtian Li$^{1}$, Xuesong Li$^{2}$ and Xiao Li$^{1}$
\thanks{$^{1}$Jie Hong, Tingtian Li, and Xiao Li are with CI$^{3}$ Lab, Department of Civil Engineering, The University of Hong Kong, Hong Kong SAR, China
        {\tt\small jiehong@hku.hk, tingtian@hku.hk, shell.x.li@hku.hk}
        }%
\thanks{$^{2}$Xuesong Li is with Commonwealth Scientific and Industrial Research Organisation (CSIRO), Canberra, ACT, Australia
        {\tt\small xuesong.li@csiro.au}}%
}

\begin{document}

\maketitle
\thispagestyle{empty}
\pagestyle{empty}

\input{sec/0_abstract}    
\input{sec/1_intro}
\input{sec/2_related_work}
\input{sec/3_method}
\input{sec/4_experiments}
\input{sec/5_conclusion}

\bibliographystyle{IEEEtran}
\bibliography{IEEEfull}

\end{document}

%% file: sec/0_abstract.tex
\begin{abstract}
Depth estimation from thermal images is highly valuable for robotic applications in adverse conditions, such as nighttime and rainy weather. Recent studies have sought to transfer knowledge from RGB-based foundation models to thermal modalities, yet the rich hierarchical representations these models encode remain underutilized. To address this limitation, we propose RGB-HS, a novel framework for thermal-image depth estimation that leverages hierarchical supervision from an RGB-based foundation model. Specifically, we first replace the baseline thermal encoder with a foundational model and introduce a parallel RGB branch that also employs a foundational model as an encoder of the same architecture, taking RGB images as input. The alignment is then performed across multiple levels between the tokens of the two encoders, allowing the thermal student branch to capture both structural precision and semantic abstraction from the RGB teacher branch. Furthermore, we introduce verification to refine the alignment process by weighting tokens from the RGB branch based on RGB image quality. Extensive experiments on the popular benchmark demonstrate that RGB-HS achieves competitive performance and more effectively exploits the representational capacity of RGB-based foundation models for depth estimation on thermal images.
\end{abstract}

%% file: sec/1_intro.tex
\section{Introduction}
\label{sec:intro}
Depth estimation is an important task in computer vision, with applications in robotic systems for real-world scenarios~\cite {wang2019pseudo, chen2019framework, wofk2019fastdepth, li2020real, yu2022multiseam}. While significant progress has been achieved in predicting depth from RGB images~\cite{eigen2014depth, bhat2021adabins, yang2024depth1}, these methods typically rely on the RGB modality and struggle under adverse conditions such as nighttime or rain~\cite{shin2023deep}. In contrast, thermal imaging provides robust perception in low-illumination or visually degraded environments by capturing scene structure through infrared radiation. However, estimating depth directly from thermal images remains challenging due to the inherent sparsity of texture and the limited availability of thermal data.

\begin{figure}[t]
\centering
\includegraphics[width=1.0\linewidth]{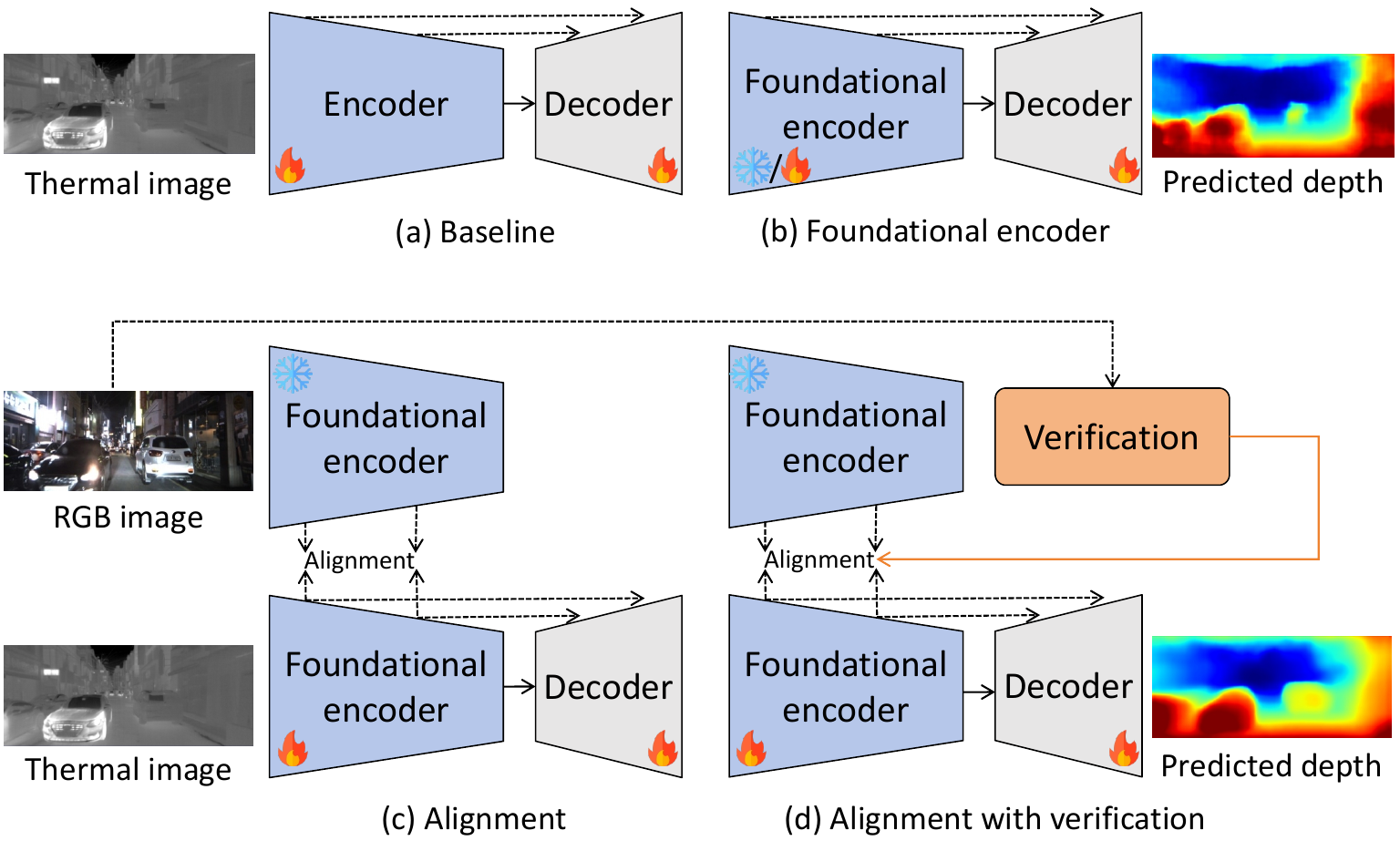}
\caption{Frameworks for depth estimation on thermal images. (a) The baseline model adopts an encoder–decoder architecture to predict depth from thermal input. (b) The baseline encoder is replaced with a powerful foundation encoder for enhanced representation learning. (c) The alignment is introduced to transfer knowledge from the RGB modality. (d) The verification is incorporated to highlight high-quality RGB data, thereby ensuring more reliable alignment.}
\label{fig:intro}
\vspace{-12pt}
\end{figure}

Recent advances in RGB-based vision foundation models~\cite{kirillov2023segment, simeoni2025dinov3, han2025enhancing} have opened new opportunities for cross-modal learning in thermal-image depth estimation. Large-scale pre-trained models, such as DINO~\cite{caron2021emerging, oquab2024dinov2, simeoni2025dinov3}, encode rich representations that generalize across diverse visual domains. Several studies have attempted to exploit these models for depth estimation on thermal images~\cite{fan2024generalizable, zuo2025monother}. For instance, GTDE~\cite{fan2024generalizable} employs DINOv2 features to enable zero-shot generalization, while RGB-MDE~\cite{zuo2025monother} distills knowledge from an RGB–depth branch to guide thermal prediction. Despite their progress, these approaches face two limitations: they do not fully leverage the RGB modality or the multi-level, structural, and semantic information embedded in RGB-based foundation models, and they treat all RGB supervision equally, overlooking the varying quality of RGB data and feature representations across different environmental conditions.

To overcome these limitations, we propose \textbf{RGB-HS}, a framework for depth estimation on thermal images via \textbf{H}ierarchical \textbf{S}upervision from RGB-based foundation models. The key idea is to transfer structural and semantic knowledge from a frozen RGB teacher encoder to a thermal student encoder through multi-level alignment, while adaptively verifying the quality of such alignment. Specifically, as illustrated in Figure~\ref{fig:intro}, RGB-HS replaces the baseline encoder with a pre-trained foundational encoder and introduces a parallel RGB branch that shares the same architecture as the thermal branch. We then design an \textit{alignment} mechanism that aligns thermal and RGB features across both map-level and latent-level spaces, enabling the thermal encoder to capture structural precision and semantic abstraction simultaneously. Furthermore, we propose a \textit{verification} module that evaluates the confidence of the RGB supervision at the image level, ensuring that unreliable RGB signals are down-weighted during knowledge transfer. In summary, our main contributions are summarized as follows:
\begin{itemize}
    \item We propose RGB-HS, a framework that leverages hierarchical supervision (\textit{i.e.}, verification-guided alignment) from an RGB-based foundation model to improve depth estimation for thermal images.
    \item For the proposed hierarchical supervision, we introduce an alignment mechanism that transfers structural and semantic cues from
    the map and latent levels. We also design a verification strategy that adaptively filters unreliable RGB supervision, thereby improving alignment quality. 
    \item Extensive experiments on the Multi-Spectral Stereo (MS$^2$) dataset~\cite{shin2023deep} demonstrate the effectiveness of RGB-HS in learning from RGB modality.
\end{itemize}

%% file: sec/2_related_work.tex
\section{Related Works}
\label{sec:related_work}
\subsubsection{Depth Estimation on RGB Images}
Depth estimation from RGB images has been extensively studied~\cite{li2015depth, yin2019enforcing, yang2023gedepth, shao2023nddepth, li2024binsformer}. The first deep learning approach for this task was introduced by Eigen \textit{et al.}~\cite{eigen2014depth}. Subsequent works have sought to enhance the prediction. For instance, Xian \textit{et al.}~\cite{xian2020structure} proposed a structure-guided ranking loss with edge- and instance-guided sampling to emphasize geometric details. AdaBins~\cite{bhat2021adabins} introduced a transformer-based architecture that adaptively partitions the depth range into image-specific bins. NeWCRF~\cite{yuan2022neural} formulated fully-connected conditional random fields (CRFs) within local windows, making them computationally feasible, and integrated them with multi-head attention for monocular depth estimation.

Recently, with the rise of RGB-based foundation models, researchers have explored adapting large pre-trained vision models for depth prediction~\cite{ke2024repurposing, yang2024depth1, yang2024depth2, simeoni2025dinov3}. Marigold~\cite{ke2024repurposing} repurposes a pre-trained Stable Diffusion v2 model~\cite{rombach2022high} via efficient fine-tuning, enabling monocular depth estimation by leveraging rich visual priors. DepthAnything~\cite{yang2024depth1} trains a foundation model using large-scale unlabeled RGB images with strong data perturbations, while DepthAnythingV2~\cite{yang2024depth2} replaces labeled data with synthetic samples to further enhance generalization. Both methods adopt DINOv2~\cite{oquab2024dinov2} as their backbone. More recently, DINOv3~\cite{simeoni2025dinov3} has been repurposed for monocular depth estimation, achieving state-of-the-art (SOTA) performance. Despite these advances, depth estimation from thermal images remains relatively underexplored.

\subsubsection{Depth Estimation on Thermal Images}
In contrast to RGB-based research, depth estimation on thermal images has only recently gained attention~\cite{shin2023deep, fan2024generalizable, zuo2025monother}. 
MTN~\cite{kim2018multispectral}, UDE~\cite{lu2021alternative}, AMA~\cite{shin2023self}, and MSCRF~\cite{shin2023deep} are the very first deep learning frameworks capable of predicting depth from either a single thermal image or a stereo thermal pair. In this work, we adopt MSCRF~\cite{shin2023deep} as the baseline.
GTDE~\cite{fan2024generalizable} proposes a two-stage strategy that leverages DINOv2~\cite{oquab2024dinov2} to achieve impressive zero-shot generalization across multiple thermal datasets. RGB-MDE~\cite{zuo2025monother} introduces confidence-aware knowledge distillation from an RGB–depth branch and employs DINOv2 as its backbone. ThermoStereoRT~\cite{hu2025thermostereort} also applies knowledge distillation for the task. Beyond depth estimation, RGB-thermal fusion has been explored in tasks such as semantic segmentation~\cite{zhang2021abmdrnet}, object tracking~\cite{zhang2023efficient, li2023efficient}, and object detection~\cite{do2024d3t, zhou2025m}.

Although GTDE~\cite{fan2024generalizable} utilizes intermediate tokens from a foundation model, it overlooks the potential of the RGB modality itself. RGB-MDE~\cite{zuo2025monother} addresses this issue by transferring knowledge from RGB images, but it neglects the information encoded within the intermediate 
tokens of the foundation model. Moreover, RGB-MDE or ThermoStereoRT does not identify and exclude low-quality RGB samples during knowledge transfer. These limitations motivate our proposed RGB-HS, which leverages hierarchical supervision to exploit both the feature- and data-level advantages of the RGB modality.

%% file: sec/3_method.tex
\section{Method}
\label{sec:method}
In this section, we propose RGB-HS, a depth estimation framework for thermal images that leverages hierarchical supervision from an RGB-based foundation model. As shown in Figure~\ref{fig:intro}, RGB-HS consists of three components: a foundation encoder, alignment, and verification.

\begin{figure*}[t]
\centering
\includegraphics[width=1.0\linewidth]{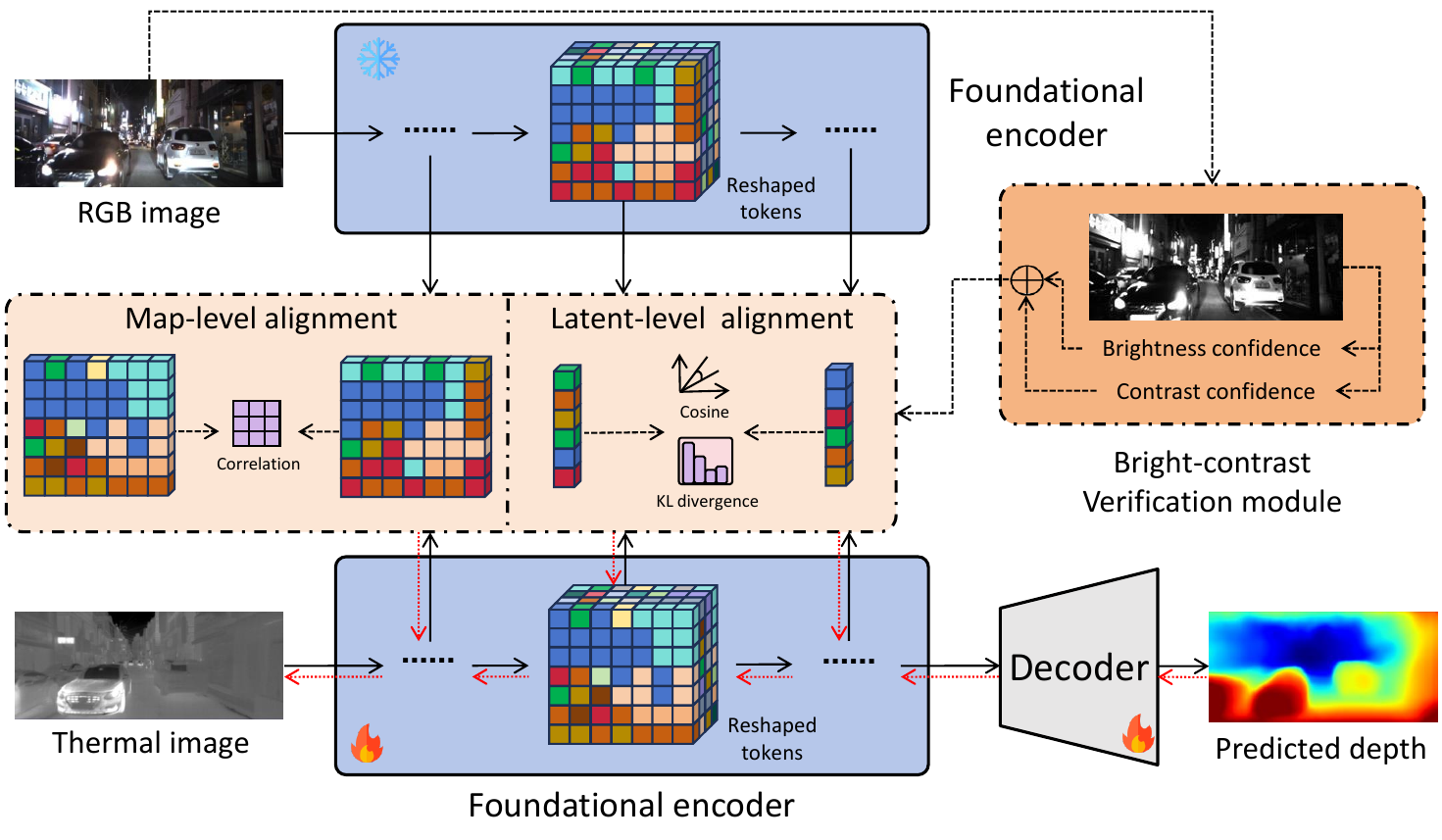}
\vspace{-18pt}
\caption{Overview of the proposed RGB-HS framework. Built upon an encoder–decoder architecture, 
RGB-HS introduces an RGB-modality branch that transfers comprehensive, reliable knowledge through alignment and verification modules.
Note that the RGB branch, alignment module, and verification module will not be used during inference.
}
\vspace{-12pt}
\label{fig:framework}
\end{figure*}

\subsection{Foundational Encoder}
In this work, we use MSCRF~\cite{shin2023deep} as the baseline, the first deep-learning framework for thermal-image depth estimation. We replace its encoder with an RGB-based foundation model (see Figure~\ref{fig:intro} (b)). MSCRF originally uses a Swin-L~\cite{liu2021swin} encoder trained from scratch. In RGB-HS, we adopt DINOv3~\cite{simeoni2025dinov3}, which achieves state-of-the-art results in RGB-based depth estimation. Specifically, we load a pre-trained ViT-B/16 backbone. The strong representation capability of DINOv3 may provide structural and semantic priors that are beneficial for depth estimation in thermal images.

\subsection{Hierarchical Supervision: Alignment}
Depth estimation from thermal images is inherently challenging due to the lack of texture and color information provided by the RGB sensor. Thermal imagery captures only temperature contrast, leading to ambiguous object boundaries and weak spatial gradients.
To better transfer knowledge from the RGB modality and make this transfer reliable, as shown in Figure~\ref{fig:intro} (c) and (d), we propose a cross-modal framework that transfers the representational power of an RGB-based foundation model (\textit{i.e.,} DINOv3~\cite{simeoni2025dinov3}) to a thermal encoder-decoder network.

More details of our method are depicted in Figure~\ref{fig:framework}. The proposed RGB-HS follows a teacher-student paradigm, where the frozen RGB teacher branch provides rich information supervision, and the thermal student branch reproduces representations for depth prediction by learning from the RGB-branch encoder. We introduce a two-level hierarchical alignment scheme: 1) Map-level alignment refines local structure by ensuring spatial correspondence, and 2) Latent-level alignment aligns global, semantic relationships between the RGB and thermal modalities.
Map-level alignment employs a correlation loss that enforces structural consistency. The latent-level alignment employs cosine similarity and Kullback–Leibler (KL) divergence loss, which enforces directional and distributional consistency.

Given a paired RGB-thermal image pair $(\mathbf{I}_{\text{RGB}}, \mathbf{I}_{\text{THR}})$, the RGB and thermal encoders extract hierarchical features:
$\mathbf{F}_{\text{RGB}} = f_{\theta_{\text{RGB}}}(\mathbf{I}_{\text{RGB}})$, $\mathbf{F}_{\text{THR}} = f_{\theta_{\text{THR}}}(\mathbf{I}_{\text{THR}})$
where $f_{\theta_{\text{RGB}}}$ and $f_{\theta_{\text{THR}}}$ denote the RGB and thermal branch encoders, respectively.
Each encoder produces multi-level features $\mathbf{F} = \{\mathbf{F}_{l}\}_{l=1}^{L}$ from its $L$ layers. For the $l$th layer, the DINOv3 transformer~\cite{simeoni2025dinov3} outputs a sequence of tokens. We extract only the spatial patch tokens $\mathbf{T}_l \in \mathbb{R}^{N_l \times C_l}$, excluding class and register tokens, where $N_l = H_l \times W_l$ is the number of spatial positions. To preserve the 2D spatial structure of the original image, these tokens are reshaped into spatial feature maps as follows:
\begin{equation}
\mathbf{F}_l = \text{Reshape}(\mathbf{T}_l) \in \mathbb{R}^{C_l \times H_l \times W_l},
\end{equation}

The decoder $g_{\phi}$ then processes the thermal features to estimate the dense depth map:
$\hat{\mathbf{D}} = g_{\phi}(\mathbf{F}_{\text{THR},1}, \mathbf{F}_{\text{THR},2}, ..., \mathbf{F}_{\text{THR},L})$,
where $\hat{\mathbf{D}}$ denotes the predicted dense depth map. 
Parallel to the thermal branch giving $\mathbf{F}_{\text{THR}}$, RGB-branch encoder $f_{\theta_{\text{RGB}}}(.)$ produces corresponding feature maps $\{\mathbf{F}_{\text{RGB},1}, \mathbf{F}_{\text{RGB},2}, \ldots, \mathbf{F}_{\text{RGB},L}\}$.  

\subsubsection{Map-level Alignment}
For depth estimation, it is natural to focus on fine-grained details such as object edges or depth discontinuities. To enhance such details in features, we introduce map-level alignment, which aims to transfer rich structural correspondence from the teacher to the student feature maps 
(reshaped tokens).
We then compute the localized alignment term: correlation alignment loss. 

\noindent \textbf{Correlation Loss.} The mismatch between RGB and thermal images exists. To avoid pixel-wise alignment but still learn local details, we align the spatial dependency of two feature maps from two modalities:
\begin{equation}
\mathcal{L}_{\text{corr}} =
\frac{1}{C} \sum_{c=1}^{C}
\big\|
\mathbf{f}_{\text{THR},c}^{\top}\mathbf{f}_{\text{THR},c} -
\mathbf{f}_{\text{RGB},c}^{\top}\mathbf{f}_{\text{RGB},c}
\big\|_{2}^{2},
\end{equation}
where $\mathbf{f}_{\text{THR},c}$ and $\mathbf{f}_{\text{RGB},c}$ are the single feature maps in $\mathbf{F}_{\text{THR}}$ or $\mathbf{F}_{\text{RGB}}$ at channel $c$. 
The map-level alignment loss is defined as:
\begin{equation}
\mathcal{L}_{\text{map}} =
\lambda^{l}\mathcal{L}_{\text{corr}},
\end{equation}\label{eqn:map_loss}
where $\lambda^{l}$ is the coefficient. 
The map-level loss $\mathcal{L}_{\text{map}}$ sharpens the structure of the feature maps learned by the thermal student branch relative to that of the RGB teacher branch.

\subsubsection{Latent-level Alignment}
While the map-level alignment promotes learning local structures, it overlooks the global semantic consistency.
Hence, we propose the latent-level alignment, which focuses on transferring high-level semantic representations from the teacher to the student.  
Here, we aim to align the global latent distributions that encode scene-wide context or semantic cues, rather than spatial pixels. Each feature map is first aggregated via spatial-wise max-pooling to form a global latent representation:
\begin{equation}
\bar{\mathbf{f}}_{\text{THR}} = \text{MaxPool}(\mathbf{F}_{\text{THR}}), \quad
\bar{\mathbf{f}}_{\text{RGB}} = \text{MaxPool}(\mathbf{F}_{\text{RGB}}),
\end{equation}
where $\bar{\mathbf{f}} \in \mathbb{R}^{C}$ summarizes the overall scene of each feature map at channel $c$. Here, we use cosine similarity loss and the KL divergence loss. The KL divergence loss helps align the probabilistic distributions of feature activations across channels, guiding the student to mimic the teacher's allocation of semantic emphasis. 

\noindent \textbf{Cosine Similarity Loss.} Similar to building the map-level loss, we then enforce latent-level alignment between the thermal and RGB embeddings using cosine similarity loss: 
\begin{equation}
\mathcal{L}_{\text{cos}}^{g} =
1 - 
\frac{\langle \bar{\mathbf{f}}_{\text{THR}}, \bar{\mathbf{f}}_{\text{RGB}} \rangle}
{\| \bar{\mathbf{f}}_{\text{THR}} \|_2 \, \| \bar{\mathbf{f}}_{\text{RGB}} \|_2},
\end{equation}
where the cosine similarity loss $\mathcal{L}_{\text{cos}}^{g}$ encourages alignment in the latent semantic space, forcing the student’s global feature vectors to lie in similar directions to those of the teacher.

\noindent \textbf{KL Divergence Loss.} To enhance the student’s sensitivity to distributional cues from the teacher, we also have KL divergence loss as follows:
\begin{equation}
\mathcal{L}_{\text{KL}}^{g} =
\text{KL}\!\left(
\text{Softmax}\!\left(\frac{\bar{\mathbf{f}}_{\text{RGB}}}{T}\right)
\,\Big\|\,
\text{Softmax}\!\left(\frac{\bar{\mathbf{f}}_{\text{THR}}}{T}\right)
\right), 
\end{equation}
where $\text{KL}(.\|.)$ computes the KL divergence between embeddings and $T$ controls the softness of the probability distribution. This term could align the channel-wise probability distributions, ensuring the student captures a similar distributional knowledge from the teacher.
The objective of latent-level alignment is defined as follows:
\begin{equation}
\mathcal{L}_{\text{latent}} =
\lambda^{g}_{1}\mathcal{L}_{\text{cos}}^{g}
+ \lambda^{g}_{2}\mathcal{L}_{\text{KL}}^{g}, 
\end{equation}\label{eqn:latent_loss}
where $\lambda^{g}_{1}$ and $\lambda^{g}_{2}$ are coefficients. By transferring latent-level information, the thermal student might gain global semantic awareness, thereby improving depth estimation in textureless or low-contrast thermal scenes.

\subsection{Hierarchical Supervision: Verification}
A critical challenge in aligning RGB-thermal features arises from the heterogeneous reliability of the RGB modality across varying scene conditions. While the frozen RGB teacher branch yields robust semantic representations under normal conditions, its features become less informative in low-light or adverse weather conditions. Thus, aligning thermal features to RGB features in all scenarios risks suppressing the thermal modality's unique capabilities. To address this, we propose an adaptive confidence mechanism that dynamically modulates the alignment strength based on confidence signals.

\noindent \textbf{Brightness–contrast Confidence.}
To assess the perceptual quality of each RGB input, we introduce a simple yet effective brightness–contrast confidence measure. Given an RGB image $\mathbf{I}_{\text{RGB}} \in \mathbb{R}^{3\times H\times W}$ with intensity values in $[0, 255]$, we first normalize it to $[0,1]$ and compute its luminance map as follows:
\begin{equation}
\mathbf{Y} = 0.2126\mathbf{R} + 0.7152\mathbf{G} + 0.0722\mathbf{B},
\end{equation}
where $\mathbf{Y}$, $\mathbf{R}$, $\mathbf{G}$, and $\mathbf{B} \in \mathbb{R}^{H\times W}$ denote the red, green, and blue channels after normalization, respectively. The average luminance and its standard deviation are used to describe the image's brightness and contrast:
\begin{equation}
C_{\text{BT}} = \frac{1}{H W} \sum_{i,j} \mathbf{Y}_{i,j},
\end{equation}
\begin{equation}
C_{\text{CT}} = \sqrt{\frac{1}{H W}\sum_{i,j} (\mathbf{Y}_{i,j} - C_{\text{BT}})^2},
\end{equation}
where $C_{\text{BT}}, C_{\text{CT}} \in \mathbb{R}$. To ensure comparability across images, $C_{\text{CT}}$ is then normalized into $[0,1]$.
To reflect both image exposure quality and contrast sharpness, the brightness–contrast confidence is the average of these two terms: 
\begin{equation}
C_{\mathrm{RGB}} = \frac{C_{\text{BT}} + C_{\text{CT}}}{2},
\end{equation}
where $C_{\mathrm{RGB}} \in \mathbb{R}$. Intuitively, $C_{\mathrm{RGB}}$ is high when the image is neither underexposed nor overexposed and exhibits adequate contrast. It can serve as a simple and interpretable probabilistic indicator of the visual quality of $\mathbf{I}_{\text{RGB}}$. We then have $\mathbf{F}_{\text{THR}} \gets C_{\mathrm{RGB}} \cdot \mathbf{F}_{\text{THR}}$ as verified features for the alignment loss calculation.

\subsection{Overall Loss}
The complete training objective integrates supervised depth learning of the baseline with both map- and latent-level alignment terms:
\begin{equation}
\mathcal{L}_{\text{total}} =
\mathcal{L}_{\text{baseline}}
+ \frac{1}{NL} \big( \sum_{n=1}^{N} \sum_{l=1}^{L} \mathcal{L}_{\text{map}}^{(n,l)}
+ \sum_{n=1}^{N} \sum_{l=1}^{L} \mathcal{L}_{\text{latent}}^{(n,l)} \big),
\end{equation} \label{eqn:final_loss}
where $\mathcal{L}_{\text{baseline}}$ denotes the supervised regression loss of the baseline, $L$ is the number of encoder layers, $N$ is the batchsize. The hierarchical design allows the thermal branch to learn both structural precision and semantic abstraction from the RGB branch.
At inference time, only the thermal encoder–decoder is retained, eliminating any dependency on RGB input.

%% file: sec/4_experiments.tex
\section{Experiments}
\label{sec:exp}
\begin{table*}[t]
\centering
\caption{Depth estimation results for thermal images on the MS$^2$~\cite{shin2023deep} dataset. The best two numbers in ``Avg'' are in bold. The best numbers are in \textcolor{red}{red}, and the second-best are in \textcolor{blue}{blue}. The results of DORN~\cite{fu2018deep}, BTS~\cite{lee2019big}, AdaBins~\cite{bhat2021adabins}, NeWCRF~\cite{yuan2022neural}, and MSCRF~\cite{shin2023deep} are provided in \cite{shin2023deep}. Note that RGB-HS does not use the RGB image as input during inference in all experiments.}
\label{tab:all_results_mono_stereo}
\resizebox{0.98\textwidth}{!}{
\begin{tabular}{llccccccc}
\toprule
\multirow{2}{*}{\textbf{Method}} & \multirow{2}{*}{\textbf{TestSet}} & \multicolumn{4}{c}{\textbf{Error $\downarrow$}} & \multicolumn{3}{c}{\textbf{Accuracy $\uparrow$}} \\
\cmidrule(lr){3-6} \cmidrule(lr){7-9}
& & \textbf{AbsRel} & \textbf{SqRel} & \textbf{RMSE} & \textbf{RMSElog} & $\delta < 1.25$ & $\delta < 1.25^2$ & $\delta < 1.25^3$ \\
\midrule

\multirow{4}{*}{\textbf{DORN}~\cite{fu2018deep}} 
& \textbf{Day}  & 0.144 & 1.288 & 5.483 & 0.230 & 0.856 & 0.941 & 0.970 \\
& \textbf{Night} & 0.136 & 1.136 & 5.290 & 0.212 & 0.863 & 0.950 & 0.976 \\
& \textbf{Rain}  & 0.180 & 1.934 & 6.735 & 0.276 & 0.781 & 0.910 & 0.955 \\
\cline{2-9}
& \textbf{Avg}  & 0.151 & 1.419 & 5.776 & 0.237 & 0.837 & 0.935 & 0.968 \\
\hline

\multirow{4}{*}{\textbf{BTS}~\cite{lee2019big}} 
& \textbf{Day}  & 0.122 & 0.905 & 4.923 & 0.198 & 0.857 & 0.951 & 0.980 \\
& \textbf{Night} & 0.114 & 0.798 & 4.701 & 0.184 & 0.870 & 0.959 & 0.984 \\
& \textbf{Rain}  & 0.157 & 1.395 & 6.053 & 0.243 & 0.791 & 0.926 & 0.969 \\
\cline{2-9}
& \textbf{Avg}  & 0.129 & 1.008 & 5.169 & 0.206 & 0.843 & 0.947 & 0.978 \\
\hline

\multirow{4}{*}{\textbf{AdaBins}~\cite{bhat2021adabins}} 
& \textbf{Day}  & 0.129 & 0.976 & 5.108 & 0.205 & 0.847 & 0.947 & 0.979 \\
& \textbf{Night} & 0.119 & 0.822 & 4.749 & 0.187 & 0.864 & 0.958 & 0.984 \\
& \textbf{Rain}  & 0.168 & 1.545 & 6.336 & 0.254 & 0.771 & 0.918 & 0.965 \\
\cline{2-9}
& \textbf{Avg}  & 0.137 & 1.084 & 5.330 & 0.212 & 0.831 & 0.943 & 0.977 \\
\hline

\multirow{4}{*}{\textbf{NeWCRF}~\cite{yuan2022neural}} 
& \textbf{Day}  & 0.120 & 0.864 & 4.852 & 0.195 & 0.858 & 0.952 & 0.982 \\
& \textbf{Night} & 0.112 & 0.755 & 4.594 & 0.179 & 0.875 & 0.961 & 0.985 \\
& \textbf{Rain}  & 0.155 & 1.352 & 5.956 & 0.240 & 0.795 & 0.929 & 0.970 \\
\cline{2-9}
& \textbf{Avg}  & 0.127  & 0.965 & 5.077 & 0.202 & 0.846 & 0.949 & 0.980 \\
\hline

\multirow{4}{*}{\textbf{MSCRF (Mono)}~\cite{shin2023deep}} 
& \textbf{Day}  & 0.115 & 0.983 & 4.895 & 0.201 & 0.882 & 0.952 & 0.977 \\
& \textbf{Night} & 0.107 & 0.850 & 4.658 & 0.185 & 0.894 & 0.961 & 0.981 \\
& \textbf{Rain}  & 0.152 & 1.567 & 6.020 & 0.247 & 0.822 & 0.928 & 0.964 \\
\cline{2-9}
& \textbf{Avg}  & 0.123 & 1.103 & 5.134 & 0.208 & 0.869 & 0.948 & 0.975 \\
\hline

\multirow{4}{*}{\textbf{MSCRF (Stereo)}~\cite{shin2023deep}} 
& \textbf{Day}   & 0.113 & 0.948 & 4.852 & 0.200 & 0.884 & 0.953 & 0.977 \\
& \textbf{Night} & 0.105 & 0.811 & 4.584 & 0.183 & 0.896 & 0.961 & 0.981 \\
& \textbf{Rain}  & 0.149 & 1.499 & 5.940 & 0.245 & 0.826 & 0.929 & 0.965 \\
\cline{2-9}
& \textbf{Avg}   & 0.120 & 1.057 & 5.068 & 0.207 & \textcolor{blue}{\textbf{0.872}} & 0.949 & 0.975 \\
\hline

\multirow{4}{*}{\textbf{RGB-HS (Mono)}} 
& \textbf{Day}   & 0.112  & 0.616  & 3.750  & 0.151  & 0.868  & 0.974  & 0.994 \\   
& \textbf{Night} & 0.099  & 0.452  & 3.223  & 0.131  & 0.899  & 0.985  & 0.997 \\
& \textbf{Rain}  & 0.141  & 0.883  & 4.446  & 0.181  & 0.818  & 0.962  & 0.990 \\
\cline{2-9}
& \textbf{Avg}   & \textcolor{blue}{\textbf{0.117}}  & \textcolor{blue}{\textbf{0.650}}  & \textcolor{blue}{\textbf{3.806}}  & \textcolor{blue}{\textbf{0.154}}  & 0.862  & \textcolor{blue}{\textbf{0.974}}  & \textcolor{blue}{\textbf{0.994}} \\
\hline

\multirow{4}{*}{\textbf{RGB-HS (Stereo)}} 
& \textbf{Day}   & 0.098  & 0.517  & 3.434  & 0.128  & 0.894  & 0.981  & 0.997 \\
& \textbf{Night} & 0.093  & 0.404  & 3.038  & 0.120  & 0.912  & 0.989  & 0.998 \\
& \textbf{Rain}  & 0.124  & 0.797  & 4.172  & 0.156  & 0.856  & 0.973  & 0.993 \\
\cline{2-9}
& \textbf{Avg}   & \textcolor{red}{\textbf{0.105}}  & \textcolor{red}{\textbf{0.572}}  & \textcolor{red}{\textbf{3.548}}  & \textcolor{red}{\textbf{0.134}}  & \textcolor{red}{\textbf{0.887}}  & \textcolor{red}{\textbf{0.981}}  & \textcolor{red}{\textbf{0.996}} \\

\hline
\end{tabular}
}
\vspace{-12pt}
\end{table*}

\begin{table}[t]
\centering
\caption{Depth estimation results for thermal images on the cleaned version of the dataset MS$^2$~\cite{shin2023deep}. The best two numbers in ``Avg'' are in bold. The best numbers are in \textcolor{red}{red}, and the second-best are in \textcolor{blue}{blue}. The results of DORN~\cite{fu2018deep}, BTS~\cite{lee2019big}, AdaBins~\cite{bhat2021adabins}, NeWCRF~\cite{yuan2022neural}, Zoedepth~\cite{bhat2023zoedepth}, DepthAnything~\cite{yang2024depth1}, and RGB-MDE~\cite{zuo2025monother} are provided in \cite{zuo2025monother}.}
\label{tab:supple1}
\resizebox{0.48\textwidth}{!}{
\begin{tabular}{lccccc}
\toprule
\multirow{2}{*}{\textbf{Method}} & \multirow{2}{*}{\textbf{Modality}} 
& \multicolumn{4}{c}{\textbf{Error $\downarrow$}} \\

\cmidrule(lr){3-6}
& & \textbf{AbsRel} & \textbf{SqRel} & \textbf{RMSE} & \textbf{RMSElog} \\
\midrule

\textbf{DORN}~\cite{fu2018deep}              & \textbf{Ther} & 0.109 & 0.540 & 3.660 & 0.144 \\
\textbf{BTS}~\cite{lee2019big}               & \textbf{Ther} & 0.086 & 0.380 & 3.163 & 0.117 \\
\textbf{AdaBins}~\cite{bhat2021adabins}      & \textbf{Ther} & 0.088 & 0.377 & 3.152 & 0.119 \\
\textbf{NeWCRF}~\cite{yuan2022neural}        & \textbf{Ther} & 0.080 & 0.331 & 2.937 & 0.109 \\
\textbf{Zoedepth}~\cite{bhat2023zoedepth}    & \textbf{Ther} & 0.091 & 0.425 & 3.202 & 0.123 \\
\textbf{DepthAnything}~\cite{yang2024depth1} & \textbf{Ther} & \textcolor{blue}{\textbf{0.075}} & 0.287 & 2.719 & \textcolor{blue}{\textbf{0.103}} \\
\textbf{RGB-MDE}~\cite{zuo2025monother}      & \textbf{Ther} & \textcolor{red}{\textbf{0.072}} & \textcolor{red}{\textbf{0.275}} & \textcolor{blue}{\textbf{2.677}} & \textcolor{red}{\textbf{0.100}} \\ \midrule
\textbf{RGB-HS (Ours)}                       & \textbf{Ther} & \textcolor{red}{\textbf{0.072}} & \textcolor{blue}{\textbf{0.283}} & \textcolor{red}{\textbf{2.595}} & \textcolor{red}{\textbf{0.100}} \\
\bottomrule
\end{tabular}
}
\vspace{-12pt}
\end{table}

\begin{figure*}[t]
\centering
\includegraphics[width=1.0\linewidth]{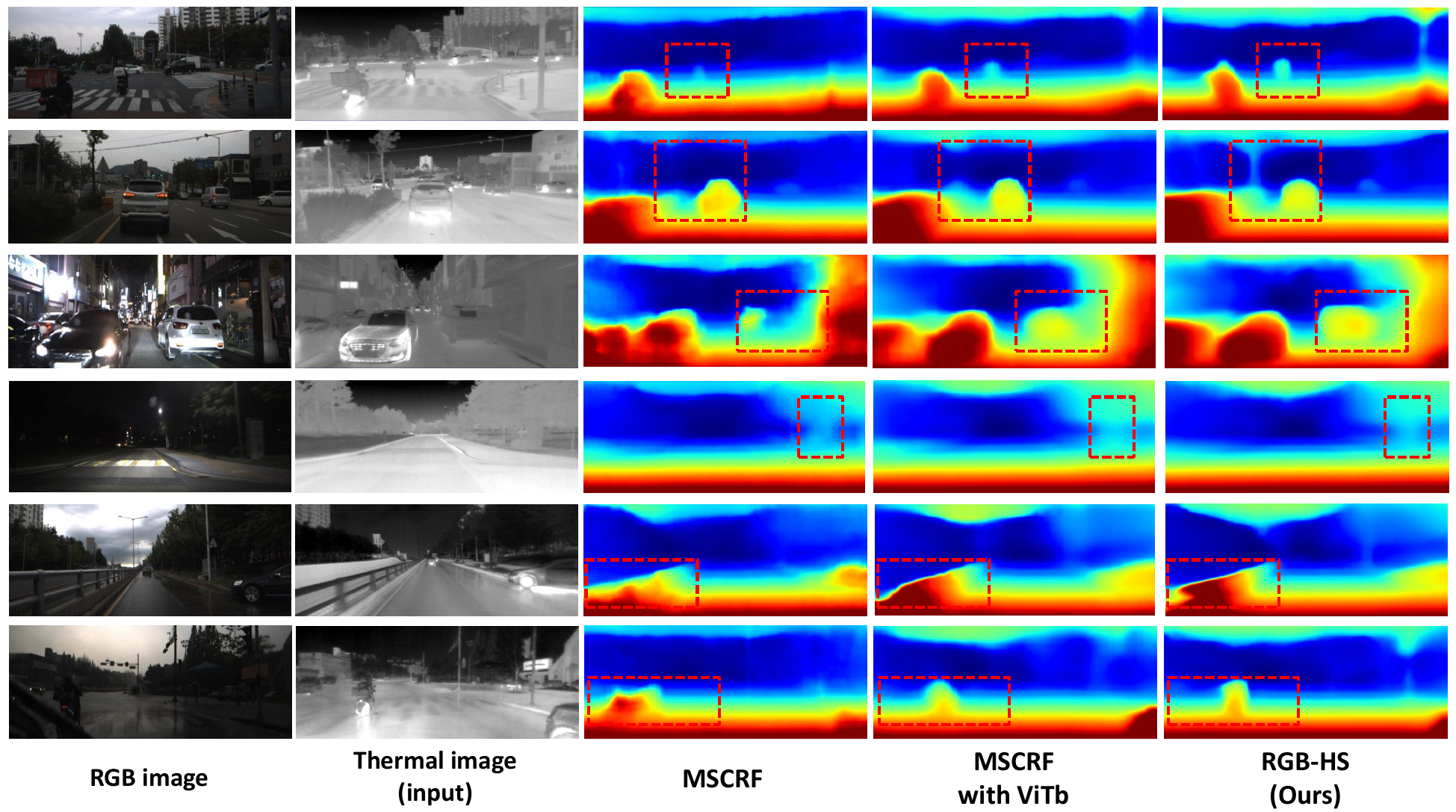}
\vspace{-12pt}
\caption{Visualizations of depth estimation results on the MS$^2$ dataset~\cite{shin2023deep}. Each row shows one example under different environmental conditions (daytime, nighttime, and rainy scenes). From left to right: RGB image (for reference), thermal input image, and predicted depth maps from MSCRF~\cite{shin2023deep}, MSCRF with DINOv3 ViT-B/16 backbone, and our RGB-HS (Ours). We highlight the representative regions with the \textcolor{red}{red} dashed boxes.}\label{fig:visual}
\vspace{-6pt}
\end{figure*}

In this section, we present the experimental results of the proposed RGB-HS framework. Experiments are conducted on the large-scale MS$^{2}$ dataset~\cite{shin2023deep}. We evaluate RGB-HS on both monocular and stereo depth estimation tasks and conduct ablation studies to assess the effectiveness of the alignment and verification mechanisms.

\subsection{Dataset}
All experiments are performed on the MS$^{2}$ dataset~\cite{shin2023deep}, a large-scale multimodal outdoor benchmark for autonomous driving. The dataset provides synchronized data from stereo RGB, near-infrared (NIR), and thermal cameras, together with stereo LiDAR ground truth. Importantly, MS$^{2}$ includes diverse illumination and weather conditions (\textit{e.g.}, nighttime and rain), making it suitable for evaluating thermal-based depth estimation. Following the official split, we use $26$K, $4$K, and $17.8$K paired samples for training, validation, and testing, respectively.

\subsection{Implementations}
The proposed RGB-HS utilized the MSCRF framework~\cite{shin2023deep} as the baseline. Hence, we adopt the same training settings and hyperparameters as those used in MSCRF. The AdamW optimizer~\cite{2019loshchilovdecoupled} is used with an initial learning rate of $1\times10^{-4}$. For the proposed alignment and verification modules, the weights of the loss terms are empirically set as $\lambda^l=0.5$, 
$\lambda^{g}_{1}=1.0$, and $\lambda^{g}_{2}=0.1$, based on validation performance. All experiments are conducted on a single NVIDIA RTX PRO 6000 GPU. Our implementation is based on PyTorch~\cite{paszke2019pytorch}.

\subsection{Depth Estimation on Thermal Images}
We first evaluate the proposed RGB-HS for monocular and stereo depth estimation using thermal images. Quantitative results on the MS$^{2}$ dataset~\cite{shin2023deep} are provided in Table~\ref{tab:all_results_mono_stereo}. We compare RGB-HS against several representative depth estimation methods on thermal images, including DORN~\cite{fu2018deep}, BTS~\cite{lee2019big}, AdaBins~\cite{bhat2021adabins}, NeWCRF~\cite{yuan2022neural}, and MSCRF~\cite{shin2023deep}. Note that RGB-HS adopts MSCRF as the baseline. Moreover, it does not use the RGB teacher branch and RGB images as input during inference.

\begin{table*}[t]
\centering
\caption{Ablation study: foundational encoder. We replace the baseline backbone with a variety of DINOv3 encoders and compare their performance. The results of stereo depth estimation are provided. The best-performing numbers are in bold.}\label{tab:results_foundational_encoder}
\resizebox{0.96\textwidth}{!}{
\begin{tabular}{llcccccccc}
\toprule
\multirow{2}{*}{\textbf{Method}} & \multirow{2}{*}{\textbf{Encoder}} & \multirow{2}{*}{\makecell{\textbf{Encoder} \\ \textbf{params}}} & \multicolumn{4}{c}{\textbf{Error $\downarrow$}} & \multicolumn{3}{c}{\textbf{Accuracy $\uparrow$}} \\
\cmidrule(lr){4-7} \cmidrule(lr){8-10}
& & & \textbf{AbsRel} & \textbf{SqRel} & \textbf{RMSE} & \textbf{RMSElog} & $\delta < 1.25$ & $\delta < 1.25^2$ & $\delta < 1.25^3$ \\
\midrule

\textbf{MSCRF}~\cite{shin2023deep}
& Swin-L    & 197M  & 0.120 & 1.057 & 5.068 & 0.207 &0.872 & 0.949 & 0.975 \\
\hline

\multirow{3}{*}{\textbf{MSCRF}~\cite{shin2023deep}} 
& ViT-S/16  & 29M   & 0.116 & 0.700 & 4.345 & 0.160 & 0.851 & 0.974 & 0.995 \\

& ViT-B/16  & 86M   & 0.111 & 0.643 & 3.826 & 0.144 & 0.873 & 0.978 & 0.995 \\

& ViT-L/16  & 300M  & 0.110 & 0.649 & 3.939 & 0.147 & 0.869 & 0.978 & 0.995 \\

\hline

\textbf{RGB-HS}
& ViT-B/16  & 86M   & \textbf{0.105}  & \textbf{0.572}  & \textbf{3.548}  & \textbf{0.134}  & \textbf{0.887}  & \textbf{0.981}  & \textbf{0.996} \\

\hline
\end{tabular}
}
\end{table*}

As shown in Table~\ref{tab:all_results_mono_stereo}, RGB-HS achieves the best performance under most evaluation metrics and testing conditions. Specifically, the proposed RGB-HS (Stereo) achieves the best average performance with an AbsRel of $0.105$, SqRel of $0.572$, RMSE/RMSElog of $3.548/0.134$, and $\delta<1.25/1.25^{2}/1.25^{3}$ accuracies of $0.887/0.981/0.996$. Even under the monocular mode, RGB-HS (Mono) outperforms previous stereo-based models. For example, compared with MSCRF (Stereo), RGB-HS (Mono) achieves an average SqRel of $0.650$ and an RMSE of $3.806$, reducing AbsRel and RMSE by $0.407$ and $1.262$, respectively. The above results demonstrate the effectiveness of leveraging hierarchical supervision from the RGB-based foundation model to enhance depth perception on thermal images. 
There is a clean version of MS$^{2}$ in which the depth of the training data is filtered~\cite{shin2025deep, zuo2025monother}.
We provide results for this clean version of the dataset in Table~\ref{tab:supple1}, in which the SOTA method RGB-MDE~\cite{zuo2025monother} is compared.

\begin{table}[t]
\centering
\caption{Ablation study: map- and latent-level alignments. The results of stereo depth estimation are provided. All models use pre-trained DINOv3 ViT-B/16~\cite{simeoni2025dinov3} as the encoder. In this case, RGB-HS is running without verification. The best-performing numbers are in bold.}\label{tab:results_distillations}
\resizebox{0.48\textwidth}{!}{
\begin{tabular}{lccccc}
\toprule
\multirow{2}{*}{\textbf{Method}} & \multirow{2}{*}{$\mathcal{L}_{\text{map}}$} & \multirow{2}{*}{$\mathcal{L}_{\text{latent}}$}  & \multicolumn{3}{c}{\textbf{Error $\downarrow$}} \\
\cmidrule(lr){4-6}
& & & \textbf{AbsRel} & \textbf{SqRel} & \textbf{RMSE} \\
\midrule

\textbf{MSCRF}~\cite{shin2023deep}
& -  & -                     &\textbf{0.111} &0.643 &3.826  \\
\hline

\multirow{3}{*}{\textbf{RGB-HS}} 
& \checkmark  & -            & 0.116 & 0.629 & 3.674 \\

& -  & \checkmark            & 0.116 & 0.631 & 3.726 \\

& \checkmark  & \checkmark   & 0.114 & \textbf{0.607} & \textbf{3.646} \\

\hline
\end{tabular}
}
\end{table}

\begin{table}[t]
\centering
\caption{Ablation study: effects of verification modules. The results of stereo depth estimation are provided. All models use pre-trained DINOv3 ViT-B/16~\cite{simeoni2025dinov3} as the encoder. The best-performing numbers are in bold.}\label{tab:results_verifications}
\resizebox{0.48\textwidth}{!}{
\begin{tabular}{lccccc}
\toprule
\multirow{2}{*}{\textbf{Method}} & \multirow{2}{*}{\textbf{Alignment}} & \multirow{2}{*}{\textbf{Verification}} & \multicolumn{3}{c}{\textbf{Error $\downarrow$}} \\
\cmidrule(lr){4-6}
& & & \textbf{AbsRel} & \textbf{SqRel} & \textbf{RMSE} \\
\midrule

\textbf{MSCRF}~\cite{shin2023deep}
& -  & -                      &0.111 &0.643 &3.826 \\
\hline

\multirow{2}{*}{\textbf{RGB-HS}} 
& \checkmark  & -             & 0.114 & 0.607 & 3.646 \\

& \checkmark  & \checkmark    & \textbf{0.105}  & \textbf{0.572}  & \textbf{3.548}  \\

\hline
\end{tabular}
}
\end{table}

The consistent gains across monocular and stereo settings demonstrate that RGB-HS generalizes well across diverse input configurations.
It is also observed from Table~\ref{tab:all_results_mono_stereo} that RGB-HS consistently maintains high performance across diverse environmental conditions. For instance, the model achieves strong accuracy at night with RMSE=$3.038$ and in rainy scenes with RMSE=$4.172$, whereas MSCRF achieves RMSE=$4.584$ and $5.940$. These results suggest that hierarchical supervision enables the model to capture robust informative cues that generalize beyond variations in illumination and weather.

We present qualitative comparisons in Figure~\ref{fig:visual}. As shown in the figure, the proposed RGB-HS generates more accurate depth maps, with more precise object boundaries than other methods. The red dashed boxes highlight representative areas, such as vehicles, cyclists, and road fences, where previous methods often produce blurred or inconsistent depth. In contrast, RGB-HS better preserves structure and maintains global depth continuity, demonstrating the effectiveness of hierarchical supervision in leveraging the knowledge of RGB-based foundation model for thermal depth estimation.
For example, as shown in the third row of Figure~\ref{fig:visual}, RGB-HS more accurately predicts the car on the right than the baselines.

\subsection{Ablation Study}
\noindent \textbf{Foundational Encoder.}
The results obtained with different foundational encoders are presented in Table~\ref{tab:results_foundational_encoder}. The baseline MSCRF~\cite{shin2023deep} utilizes Swin-L~\cite{liu2021swin} as its backbone, whereas our RGB-HS is constructed upon a pre-trained DINOv3 ViT encoder~\cite{simeoni2025dinov3}. We also replace the encoder in MSCRF with ViT models of varying scales. As shown in Table~\ref{tab:results_foundational_encoder}, to some extent, the performance improves with larger ViT capacities, suggesting that the foundation model’s strong visual priors benefit depth learning in the thermal domain. Notably, RGB-HS with ViT-B/16 achieves excellent overall performance, surpassing MSCRF by significant margins (SqRel=$0.572$, RMSE=$3.548$ vs. SqRel=$1.057$, RMSE=$5.068$) while using fewer parameters ($86$M vs. $197$M). This suggests that the hierarchical supervision from the RGB-based foundation model effectively transfers the knowledge to the thermal modality.

\noindent \textbf{Map- and Latent-level Alignments.}
We next analyze the impact of the proposed alignment module. Table~\ref{tab:results_distillations} summarizes the results when enabling or disabling the map-level loss $\mathcal{L}_{\text{map}}$ and the latent-level loss $\mathcal{L}_{\text{latent}}$. Compared to the baseline MSCRF without alignment, adding either loss alone marginally improves both error (\textit{i.e.}, AbsRel, SqRel, and RMSE) and accuracy performance. The combination of both yields the best performance across all metrics, suggesting complementary roles. The map-level alignment locally enforces structural consistency with RGB features, while the latent-level alignment provides higher-level semantic alignment, jointly enhancing the thermal encoder’s representational power.

\noindent \textbf{Effects of Verification Module.}
Here, we evaluate the contribution of the proposed verification modules. As shown in Table~\ref{tab:results_verifications}, introducing RGB-image-quality verification yields improvements over the version without verification. The results indicate that the verification mechanism acts as a quality filter, promoting the use of RGB data for alignment and enhancing the reliability of cross-modal supervision. 

\noindent \textbf{Model Complexity Comparison.}
As shown in Table~\ref{tab:method_comparison_booktabs}, the proposed RGB-HS achieves the best complexity performance in terms of model size and FLOPs. Even though RGB-HS's depth estimation performance is close to that of RGB-MDE (see Table~\ref{tab:supple1}), it significantly outperforms RGB-MDE across all complexity metrics.

\begin{table}[t]
\centering
\caption{Model complexity comparison of monocular depth estimation methods on thermal images.}
\label{tab:method_comparison_booktabs}
\resizebox{0.48\textwidth}{!}{
\begin{tabular}{lccc}
\toprule
\multirow{2}{*}{\textbf{Method}} & \multirow{2}{*}{\textbf{Model size (M) $\downarrow$}}  & \multirow{2}{*}{\makecell{\textbf{Inference time} \\ \textbf{(ms/per image)} $\downarrow$}}  & \multirow{2}{*}{\textbf{FLOPs (T) $\downarrow$}} \\
\\
\midrule
\textbf{MSCRF}~\cite{shin2023deep}        & 284   & \textbf{31.86}  & 0.32 \\
\textbf{RGB-MDE}~\cite{zuo2025monother}   & 666   & 698.22          & 0.72 \\ 
\midrule
\textbf{RGB-HS (Ours)}                    & \textbf{249}  & 44.46  & \textbf{0.23} \\
\bottomrule
\end{tabular}
}
\end{table}

%% file: sec/5_conclusion.tex
\section{Conclusion}
\label{sec:conclusion}
This work presents \textbf{RGB-HS}, a framework for depth estimation on thermal images that leverages hierarchical supervision from RGB-based foundation models. By introducing a frozen RGB teacher encoder and transferring knowledge to a thermal student encoder via \textit{alignment}, RGB-HS enables the thermal branch to learn both structural precision and semantic abstraction. Furthermore, the proposed \textit{verification} mechanism adaptively reduces the weight of unreliable RGB supervision by evaluating the confidence of image-level representation, thereby ensuring more accurate alignment. Extensive experiments on the MS$^2$ benchmark demonstrate that RGB-HS not only achieves competitive performance but also better exploits the representational capacity of RGB foundation models for thermal perception. Extending similar hierarchical supervision to other tasks, such as thermal-based semantic segmentation, detection, and tracking, could be a promising future research direction that further bridges the gap between visible- and infrared-based perception.

%% file: IEEEfull.bib
@article{li2020real,
  title={Real-time 3D object proposal generation and classification using limited processing resources},
  author={Li, Xuesong and Guivant, Jose and Khan, Subhan},
  journal={Robotics and Autonomous Systems},
  volume={130},
  pages={103557},
  year={2020},
  publisher={Elsevier}
}

@inproceedings{shin2023deep,
  title={Deep depth estimation from thermal image},
  author={Shin, Ukcheol and Park, Jinsun and Kweon, In So},
  booktitle={Proceedings of the IEEE/CVF Conference on Computer Vision and Pattern Recognition},
  pages={1043--1053},
  year={2023}
}

@article{shin2025deep,
  title={Deep Depth Estimation from Thermal Image: Dataset, Benchmark, and Challenges},
  author={Shin, Ukcheol and Park, Jinsun},
  journal={arXiv preprint arXiv:2503.22060},
  year={2025}
}

@inproceedings{fu2018deep,
  title={Deep ordinal regression network for monocular depth estimation},
  author={Fu, Huan and Gong, Mingming and Wang, Chaohui and Batmanghelich, Kayhan and Tao, Dacheng},
  booktitle={Proceedings of the IEEE conference on computer vision and pattern recognition},
  pages={2002--2011},
  year={2018}
}

@article{zuo2025monother,
  title={MonoTher-Depth: Enhancing Thermal Depth Estimation via Confidence-Aware Distillation},
  author={Zuo, Xingxing and Ranganathan, Nikhil and Lee, Connor and Gkioxari, Georgia and Chung, Soon-Jo},
  journal={IEEE Robotics and Automation Letters},
  year={2025},
  publisher={IEEE}
}

@inproceedings{fan2024generalizable,
  title={Generalizable Thermal-based Depth Estimation via Pre-trained Visual Foundation Model},
  author={Fan, Ruoyu and Zhao, Wang and Lin, Matthieu and Wang, Qi and Liu, Yong-Jin and Wang, Wenping},
  booktitle={2024 IEEE International Conference on Robotics and Automation (ICRA)},
  pages={14614--14621},
  year={2024},
  organization={IEEE}
}

@article{bhat2023zoedepth,
  title={Zoedepth: Zero-shot transfer by combining relative and metric depth},
  author={Bhat, Shariq Farooq and Birkl, Reiner and Wofk, Diana and Wonka, Peter and M{\"u}ller, Matthias},
  journal={arXiv preprint arXiv:2302.12288},
  year={2023}
}

@article{lee2019big,
  title={From big to small: Multi-scale local planar guidance for monocular depth estimation},
  author={Lee, Jin Han and Han, Myung-Kyu and Ko, Dong Wook and Suh, Il Hong},
  journal={arXiv preprint arXiv:1907.10326},
  year={2019}
}

@inproceedings{caron2021emerging,
  title={Emerging properties in self-supervised vision transformers},
  author={Caron, Mathilde and Touvron, Hugo and Misra, Ishan and J{\'e}gou, Herv{\'e} and Mairal, Julien and Bojanowski, Piotr and Joulin, Armand},
  booktitle={Proceedings of the IEEE/CVF international conference on computer vision},
  pages={9650--9660},
  year={2021}
}

@article{oquab2024dinov2,
  title={DINOv2: Learning Robust Visual Features without Supervision},
  author={Oquab, Maxime and Darcet, Timoth{\'e}e and Moutakanni, Th{\'e}o and Vo, Huy and Szafraniec, Marc and Khalidov, Vasil and Fernandez, Pierre and Haziza, Daniel and Massa, Francisco and El-Nouby, Alaaeldin and others},
  journal={Transactions on Machine Learning Research},
  year={2024}
}

@article{simeoni2025dinov3,
  title={Dinov3},
  author={Sim{\'e}oni, Oriane and Vo, Huy V and Seitzer, Maximilian and Baldassarre, Federico and Oquab, Maxime and Jose, Cijo and Khalidov, Vasil and Szafraniec, Marc and Yi, Seungeun and Ramamonjisoa, Micha{\"e}l and others},
  journal={arXiv preprint arXiv:2508.10104},
  year={2025}
}

@inproceedings{rombach2022high,
  title={High-resolution image synthesis with latent diffusion models},
  author={Rombach, Robin and Blattmann, Andreas and Lorenz, Dominik and Esser, Patrick and Ommer, Bj{\"o}rn},
  booktitle={Proceedings of the IEEE/CVF conference on computer vision and pattern recognition},
  pages={10684--10695},
  year={2022}
}

@inproceedings{kirillov2023segment,
  title={Segment anything},
  author={Kirillov, Alexander and Mintun, Eric and Ravi, Nikhila and Mao, Hanzi and Rolland, Chloe and Gustafson, Laura and Xiao, Tete and Whitehead, Spencer and Berg, Alexander C and Lo, Wan-Yen and others},
  booktitle={Proceedings of the IEEE/CVF international conference on computer vision},
  pages={4015--4026},
  year={2023}
}

@inproceedings{liu2021swin,
  title={Swin transformer: Hierarchical vision transformer using shifted windows},
  author={Liu, Ze and Lin, Yutong and Cao, Yue and Hu, Han and Wei, Yixuan and Zhang, Zheng and Lin, Stephen and Guo, Baining},
  booktitle={Proceedings of the IEEE/CVF international conference on computer vision},
  pages={10012--10022},
  year={2021}
}

@inproceedings{2019loshchilovdecoupled,
  title={Decoupled Weight Decay Regularization},
  author={Loshchilov, Ilya and Hutter, Frank},
  booktitle={International Conference on Learning Representations},
  year={2019}
}

@article{paszke2019pytorch,
  title={Pytorch: An imperative style, high-performance deep learning library},
  author={Paszke, Adam and Gross, Sam and Massa, Francisco and Lerer, Adam and Bradbury, James and Chanan, Gregory and Killeen, Trevor and Lin, Zeming and Gimelshein, Natalia and Antiga, Luca and others},
  journal={Advances in neural information processing systems},
  volume={32},
  year={2019}
}

@inproceedings{wang2019pseudo,
  title={Pseudo-lidar from visual depth estimation: Bridging the gap in 3d object detection for autonomous driving},
  author={Wang, Yan and Chao, Wei-Lun and Garg, Divyansh and Hariharan, Bharath and Campbell, Mark and Weinberger, Kilian Q},
  booktitle={Proceedings of the IEEE/CVF conference on computer vision and pattern recognition},
  pages={8445--8453},
  year={2019}
}

@inproceedings{wofk2019fastdepth,
  title={Fastdepth: Fast monocular depth estimation on embedded systems},
  author={Wofk, Diana and Ma, Fangchang and Yang, Tien-Ju and Karaman, Sertac and Sze, Vivienne},
  booktitle={2019 International Conference on Robotics and Automation (ICRA)},
  pages={6101--6108},
  year={2019},
  organization={IEEE}
}

@article{eigen2014depth,
  title={Depth map prediction from a single image using a multi-scale deep network},
  author={Eigen, David and Puhrsch, Christian and Fergus, Rob},
  journal={Advances in neural information processing systems},
  volume={27},
  year={2014}
}

@article{li2023efficient,
  title={Efficient and accurate object detection with simultaneous classification and tracking under limited computing power},
  author={Li, Xuesong and Guivant, Jose E},
  journal={IEEE Transactions on Intelligent Transportation Systems},
  volume={24},
  number={6},
  pages={5740--5751},
  year={2023},
  publisher={IEEE}
}

@inproceedings{bhat2021adabins,
  title={Adabins: Depth estimation using adaptive bins},
  author={Bhat, Shariq Farooq and Alhashim, Ibraheem and Wonka, Peter},
  booktitle={Proceedings of the IEEE/CVF conference on computer vision and pattern recognition},
  pages={4009--4018},
  year={2021}
}

@article{li2024binsformer,
  title={BinsFormer: Revisiting Adaptive Bins for Monocular Depth Estimation},
  author={Li, Zhenyu and Wang, Xuyang and Liu, Xianming and Jiang, Junjun},
  journal={IEEE Transactions on Image Processing},
  volume={33},
  pages={3964--3976},
  year={2024},
  publisher={IEEE}
}

@inproceedings{yin2019enforcing,
  title={Enforcing geometric constraints of virtual normal for depth prediction},
  author={Yin, Wei and Liu, Yifan and Shen, Chunhua and Yan, Youliang},
  booktitle={Proceedings of the IEEE/CVF international conference on computer vision},
  pages={5684--5693},
  year={2019}
}

@inproceedings{xian2020structure,
  title={Structure-guided ranking loss for single image depth prediction},
  author={Xian, Ke and Zhang, Jianming and Wang, Oliver and Mai, Long and Lin, Zhe and Cao, Zhiguo},
  booktitle={Proceedings of the IEEE/CVF Conference on Computer Vision and Pattern Recognition},
  pages={611--620},
  year={2020}
}

@inproceedings{li2015depth,
  title={Depth and surface normal estimation from monocular images using regression on deep features and hierarchical crfs},
  author={Li, Bo and Shen, Chunhua and Dai, Yuchao and Van Den Hengel, Anton and He, Mingyi},
  booktitle={Proceedings of the IEEE conference on computer vision and pattern recognition},
  pages={1119--1127},
  year={2015}
}

@inproceedings{yuan2022neural,
  title={Neural window fully-connected crfs for monocular depth estimation},
  author={Yuan, Weihao and Gu, Xiaodong and Dai, Zuozhuo and Zhu, Siyu and Tan, Ping},
  booktitle={Proceedings of the IEEE/CVF conference on computer vision and pattern recognition},
  pages={3916--3925},
  year={2022}
}

@inproceedings{yang2023gedepth,
  title={Gedepth: Ground embedding for monocular depth estimation},
  author={Yang, Xiaodong and Ma, Zhuang and Ji, Zhiyu and Ren, Zhe},
  booktitle={Proceedings of the IEEE/CVF international conference on computer vision},
  pages={12719--12727},
  year={2023}
}

@inproceedings{shao2023nddepth,
  title={Nddepth: Normal-distance assisted monocular depth estimation},
  author={Shao, Shuwei and Pei, Zhongcai and Chen, Weihai and Wu, Xingming and Li, Zhengguo},
  booktitle={Proceedings of the IEEE/CVF International Conference on Computer Vision},
  pages={7931--7940},
  year={2023}
}

@inproceedings{ke2024repurposing,
  title={Repurposing diffusion-based image generators for monocular depth estimation},
  author={Ke, Bingxin and Obukhov, Anton and Huang, Shengyu and Metzger, Nando and Daudt, Rodrigo Caye and Schindler, Konrad},
  booktitle={Proceedings of the IEEE/CVF conference on computer vision and pattern recognition},
  pages={9492--9502},
  year={2024}
}

@inproceedings{yang2024depth1,
  title={Depth anything: Unleashing the power of large-scale unlabeled data},
  author={Yang, Lihe and Kang, Bingyi and Huang, Zilong and Xu, Xiaogang and Feng, Jiashi and Zhao, Hengshuang},
  booktitle={Proceedings of the IEEE/CVF conference on computer vision and pattern recognition},
  pages={10371--10381},
  year={2024}
}

@article{yang2024depth2,
  title={Depth anything v2},
  author={Yang, Lihe and Kang, Bingyi and Huang, Zilong and Zhao, Zhen and Xu, Xiaogang and Feng, Jiashi and Zhao, Hengshuang},
  journal={Advances in Neural Information Processing Systems},
  volume={37},
  pages={21875--21911},
  year={2024}
}

@inproceedings{zhang2021abmdrnet,
  title={ABMDRNet: Adaptive-weighted bi-directional modality difference reduction network for RGB-T semantic segmentation},
  author={Zhang, Qiang and Zhao, Shenlu and Luo, Yongjiang and Zhang, Dingwen and Huang, Nianchang and Han, Jungong},
  booktitle={Proceedings of the IEEE/CVF Conference on Computer Vision and Pattern Recognition},
  pages={2633--2642},
  year={2021}
}

@inproceedings{zhang2023efficient,
  title={Efficient rgb-t tracking via cross-modality distillation},
  author={Zhang, Tianlu and Guo, Hongyuan and Jiao, Qiang and Zhang, Qiang and Han, Jungong},
  booktitle={Proceedings of the IEEE/CVF conference on computer vision and pattern recognition},
  pages={5404--5413},
  year={2023}
}

@inproceedings{do2024d3t,
  title={D3t: Distinctive dual-domain teacher zigzagging across rgb-thermal gap for domain-adaptive object detection},
  author={Do, Dinh Phat and Kim, Taehoon and Na, Jaemin and Kim, Jiwon and Lee, Keonho and Cho, Kyunghwan and Hwang, Wonjun},
  booktitle={Proceedings of the IEEE/CVF Conference on Computer Vision and Pattern Recognition},
  pages={23313--23322},
  year={2024}
}

@inproceedings{shin2023self,
  title={Self-supervised monocular depth estimation from thermal images via adversarial multi-spectral adaptation},
  author={Shin, Ukcheol and Park, Kwanyong and Lee, Byeong-Uk and Lee, Kyunghyun and Kweon, In So},
  booktitle={Proceedings of the IEEE/CVF Winter Conference on Applications of Computer Vision},
  pages={5798--5807},
  year={2023}
}

@inproceedings{lu2021alternative,
  title={An alternative of lidar in nighttime: Unsupervised depth estimation based on single thermal image},
  author={Lu, Yawen and Lu, Guoyu},
  booktitle={Proceedings of the IEEE/CVF Winter Conference on Applications of Computer Vision},
  pages={3833--3843},
  year={2021}
}

@inproceedings{kim2018multispectral,
  title={Multispectral transfer network: Unsupervised depth estimation for all-day vision},
  author={Kim, Namil and Choi, Yukyung and Hwang, Soonmin and Kweon, In So},
  booktitle={Proceedings of the AAAI Conference on Artificial Intelligence},
  volume={32},
  number={1},
  year={2018}
}

@inproceedings{hu2025thermostereort,
  title={ThermoStereoRT: Thermal Stereo Matching in Real Time via Knowledge Distillation and Attention-based Refinement},
  author={Hu, Anning and Li, Ang and Jin, Xirui and Zou, Danping},
  booktitle={2025 IEEE International Conference on Robotics and Automation (ICRA)},
  pages={3766--3772},
  year={2025},
  organization={IEEE}
}

@inproceedings{zhou2025m,
  title={M-SpecGene: Generalized Foundation Model for RGBT Multispectral Vision},
  author={Zhou, Kailai and Yang, Fuqiang and Wang, Shixian and Wen, Bihan and Zi, Chongde and Chen, Linsen and Shen, Qiu and Cao, Xun},
  booktitle={Proceedings of the IEEE/CVF International Conference on Computer Vision},
  pages={7861--7872},
  year={2025}
}

@inproceedings{chen2019framework,
  title={A framework for 3D object detection and pose estimation in unstructured environment using single shot detector and refined LineMod template matching},
  author={Chen, Shili and Hong, Jie and Liu, Xineng and Li, Jian and Zhang, Tao and Wang, Danwei and Guan, Yisheng},
  booktitle={2019 24th IEEE International Conference on Emerging Technologies and Factory Automation (ETFA)},
  pages={499--504},
  year={2019},
  organization={IEEE}
}

@article{yu2022multiseam,
  title={Multiseam tracking with a portable robotic welding system in unstructured environments},
  author={Yu, Shuangfei and Guan, Yisheng and Yang, Zhi and Liu, Chutian and Hu, Jiacheng and Hong, Jie and Zhu, Haifei and Zhang, Tao},
  journal={The International Journal of Advanced Manufacturing Technology},
  volume={122},
  number={3},
  pages={2077--2094},
  year={2022},
  publisher={Springer}
}

@inproceedings{han2025enhancing,
  title={Enhancing features in long-tailed data using large vision model},
  author={Han, Pengxiao and Ye, Changkun and Tong, Jinguang and Jiang, Cuicui and Hong, Jie and Fang, Li and Li, Xuesong},
  booktitle={2025 International Joint Conference on Neural Networks (IJCNN)},
  pages={1--9},
  year={2025},
  organization={IEEE}
}
